\documentclass[fleqn,10pt]{wlscirep}

\title{DeepWeeds: A Multiclass Weed Species Image Dataset for Deep Learning}

\author[1,*]{Alex Olsen}
\author[1]{Dmitry A. Konovalov}
\author[1]{Bronson Philippa}
\author[1]{Peter Ridd}
\author[1]{Jake C. Wood}
\author[1]{Jamie Johns}
\author[1]{Wesley Banks}
\author[1]{Benjamin Girgenti}
\author[1]{Owen Kenny}
\author[1]{James Whinney}
\author[1]{Brendan Calvert}
\author[1]{Mostafa Rahimi Azghadi}
\author[1]{Ronald D. White}

\affil[1]{College of Science and Engineering, James Cook University, Townsville, QLD, 4811, Australia}
\affil[*]{Correspondence and requests for materials should be addressed to \href{mailto:alex.olsen@jcu.edu.au}{alex.olsen@my.jcu.edu.au}}

\usepackage{colortbl}
\usepackage{subcaption}
\usepackage{bookmark}
\usepackage{multirow}
\usepackage{graphicx}
\usepackage{gensymb}

\begin{abstract}
Robotic weed control has seen increased research of late with its potential for boosting productivity in agriculture. Majority of works focus on developing robotics for croplands, ignoring the weed management problems facing rangeland stock farmers. Perhaps the greatest obstacle to widespread uptake of robotic weed control is the robust classification of weed species in their natural environment. The unparalleled successes of deep learning make it an ideal candidate for recognising various weed species in the complex rangeland environment. This work contributes the first large, public, multiclass image dataset of weed species from the Australian rangelands; allowing for the development of robust classification methods to make robotic weed control viable. The \emph{DeepWeeds} dataset consists of 17,509 labelled images of eight nationally significant weed species native to eight locations across northern Australia. This paper presents a baseline for classification performance on the dataset using the benchmark deep learning models, Inception-v3 and ResNet-50. These models achieved an average classification accuracy of 95.1\% and 95.7\%, respectively. We also demonstrate real time performance of the ResNet-50 architecture, with an average inference time of 53.4 ms per image. These strong results bode well for future field implementation of robotic weed control methods in the Australian rangelands.
\end{abstract}

\begin{document}
\flushbottom
\maketitle

\section*{Introduction}
Robotic weed control promises a step-change in agricultural productivity \cite{Gonzalez-de-Santos2017,Fernandez-Quintanilla2018}. The primary benefits of autonomous weed control systems are in reducing the labour cost while also potentially reducing herbicide usage with more efficient selective application to weed targets. Improving the efficacy of weed control would have enormous economic impact. In Australia alone, it is estimated that farmers spend AUD\$1.5 billion each year on weed control activities and lose a further \$2.5 billion in impacted agricultural production \cite{AgWhitePaper2015}. Successful development of agricultural robotics is likely to reduce these losses and improve productivity.

Research in robotic weed control has focused on what many consider to be the four core technologies: detection, mapping, guidance and control \cite{RoboticWeedControlReview2008}. Of these, detection and classification remains a significant obstacle toward commercial development and industry acceptance of robotic weed control technology \cite{RoboticWeedControlReview2008, RoboticWeedControlReview2014}. Three primary methods of detection exist that focus on different representations of the light spectrum. Varied success has been achieved using image-based \cite{2018-Visible-WheatCrops,DOSSANTOSFERREIRA2017314,2017-Visible-CerealCrops,Flavia2007,Kumar2012,Hall15,KALYONCU2015102,Lee20171, Lee2015452, Carranza-Rojas2017}, spectrum-based \cite{2018-Spectrum-CommonWeeds,2017-Spectrum-WinterRape} and spectral image-based \cite{2018-Hyperpectral-CropWeeds,2017-Hyperspectral-Corn-Weeds} methods to identify weeds from both ground and aerial photography. Spectrum and spectral image-based methods are most suitable for highly controlled, site-specific environments, such as arable croplands where spectrometers can be tailored to their environment for consistent acquisition and detection. However, the harsh and complex rangeland environment make spectral-based methods challenging to implement, while image-based methods benefit from cheaper and simpler image acquisition in varying light conditions, especially when deployed in a moving vehicle in real time \cite{MAHESH201517}. Thus for this work, we focus on image-based techniques for recognising weed species.

The automatic recognition of plants using computer vision algorithms is an important academic and practical challenge \cite{HoG2015}. One way of solving this challenge is to identify plants from their leaf images \cite{Flavia2007,Kumar2012,Hall15,KALYONCU2015102,DOSSANTOSFERREIRA2017314,Lee20171, Lee2015452, Carranza-Rojas2017}. A variety of algorithms and methods have been developed to solve this problem \cite{Waldchen2017}. Perhaps the most promising recent leaf-classification methods are based on deep learning models, such as Convolutional Neural Networks (CNN) \cite{DOSSANTOSFERREIRA2017314,Lee20171,Lee2015452, LeCun15}; which now dominate many computer vision related fields. For example, the ImageNet Large Scale Visual Recognition Challenge (ILSVRC) has been dominated by CNN variants since 2012 when a CNN \cite{Krizhevsky2012} won for the first time, and by a wide margin. This and other recent successes behoove the use of deep learning in the detection and classification of weed species.

\newpage

\begin{figure}[h!]
	\centering
	\includegraphics[width=0.88\textwidth]{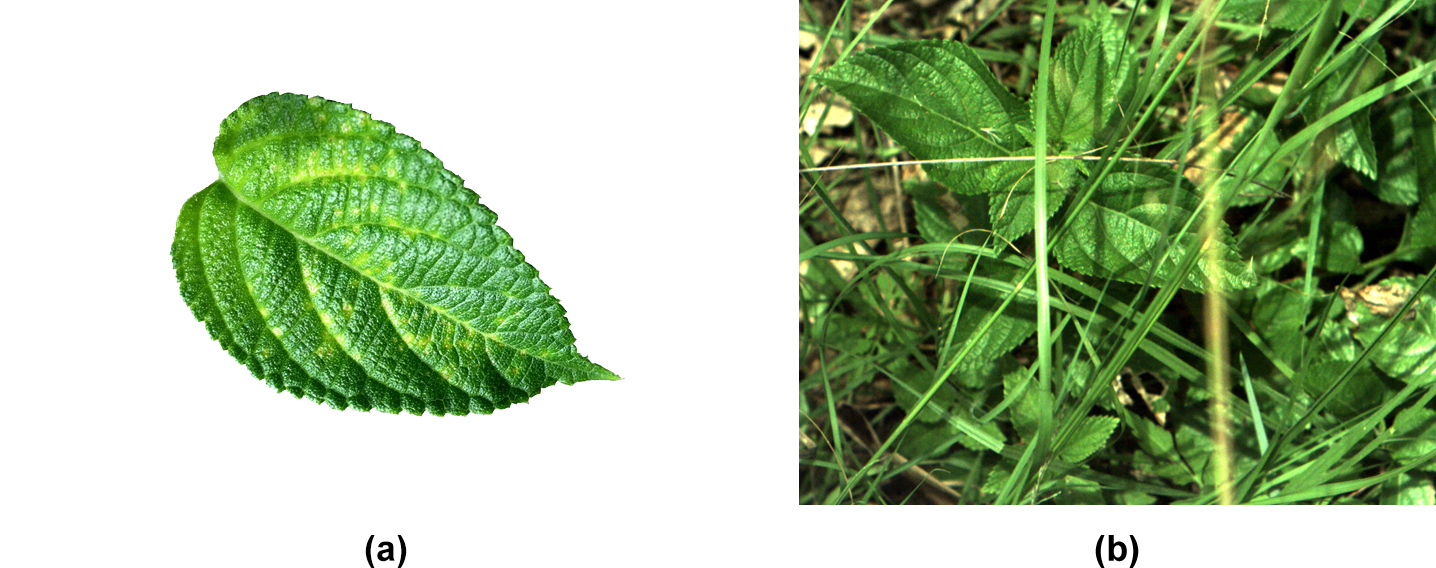}
	\caption{A comparison of the inherent difficulty associated with classifying weed species in the lab versus in the field. (a) An image of a lantana leaf taken in a controlled lab environment. (b) A sample image of lantana from the \emph{DeepWeeds} dataset, taken in situ capturing a realistic view of the entire plant.\label{LabVsField}}
\end{figure}

The performance of every machine learning model, from linear regression to CNNs, is bound by the dataset it is learning. The literature boasts many weed and plant life image datasets \cite{Flavia2007,Kumar2012,Lee2015452,HoG2015}. The annual LifeCLEF plant identification challenge \cite{CLEF2015,CLEF2016,CLEF2017} presented a 2015 dataset \cite{CLEF2015} composed of 113,205 images belonging to 41,794 observations of 1,000 species of trees, herbs and ferns. This sprawling dataset is quite unique, with most other works presenting site-specific datasets for their weeds of interest \cite{HoG2015, Flavia2007, Lee2015452}. These approaches all deliver high classification accuracy for their target datasets. However, most datasets capture their target plant life under perfect lab conditions \cite{Flavia2007,Lee2015452}. While the perfect lab conditions allow for strong theoretical classification results, deploying a classification model on a weed control robot requires an image dataset that photographs the plants under realistic environmental conditions. Figure \ref{LabVsField} illustrates the comparative difficulty of classifying species in situ.

The majority of current weed species classification methods also lean towards weed control in cropping applications \cite{PEREZORTIZ2015533, TANG2016103, 7743317, DOSSANTOSFERREIRA2017314}, where classification using machine vision is simple because the land is often flat, vegetation homogeneous and the light conditions are controlled. Classification of weeds in rangeland environments, however, has been largely ignored. Rangeland environments pose unique challenges for weed management and classification because they are remote and extensive, with rough and uneven terrain, and complex target backgrounds. Furthermore, many different species of weeds and native plants may also be present in the same area, all at varying distances from the camera, each experiencing different levels of light and shade, with some weeds being entirely hidden. To allow the classification methods deployed in this environment any chance of success, site-specific and highly variable weed species image datasets are needed.

Liaison with land care groups and property owners across northern Australia led to the selection of eight target weed species for the the collection of a large weed species image dataset; (1) chinee apple ({\itshape Ziziphus mauritiana}), (2) lantana ({\itshape Lantana camara}), (3) parkinsonia ({\itshape Parkinsonia aculeata}), (4) parthenium ({\itshape Parthenium hysterophorus}), (5) prickly acacia ({\itshape Vachellia nilotica}), (6) rubber vine ({\itshape Cryptostegia grandiflora}), (7) siam weed ({\itshape Chromolaena odorata}) and (8) snake weed ({\itshape Stachytarpheta spp.}). These species were selected because of their suitability for foliar herbicide spraying, and their notoriety for invasiveness and damaging impact to rural Australia. Five of the eight species have been targeted by the Australian Government as \emph{Weeds of National Significance} in a bid to limit their potential spread and socio-economic impacts \cite{WoNS-2012}.

In this study, we present the \emph{DeepWeeds} dataset, containing 17,509 images of eight different weed species labelled by humans. The images were collected in situ from eight rangeland environments across northern Australia. Furthermore, we train a deep learning image classifier to identify the species that are present in each image; and validate the real time performance of the classifier. We anticipate that the dataset and our classification results will inspire further research into the classification of rangeland weeds under realistic conditions as would be experienced by an autonomous weed control robot.

\section*{Methods}

\subsection*{Data Collection Platform}

The first goal of this work was the collection of a large labelled image dataset to facilitate the classification of a variety of weed species for robotic weed control. The emerging trend of deep learning for object detection and classification necessitates its use for this task. As a result, careful consideration was taken for key factors to aid the learning process. These factors include: the optical system, scene variability, dataset size, weed targets, weed locations, negative samples, image metadata and labelling.

\begin{figure}[h]
	\centering
	\includegraphics[width=0.85\textwidth]{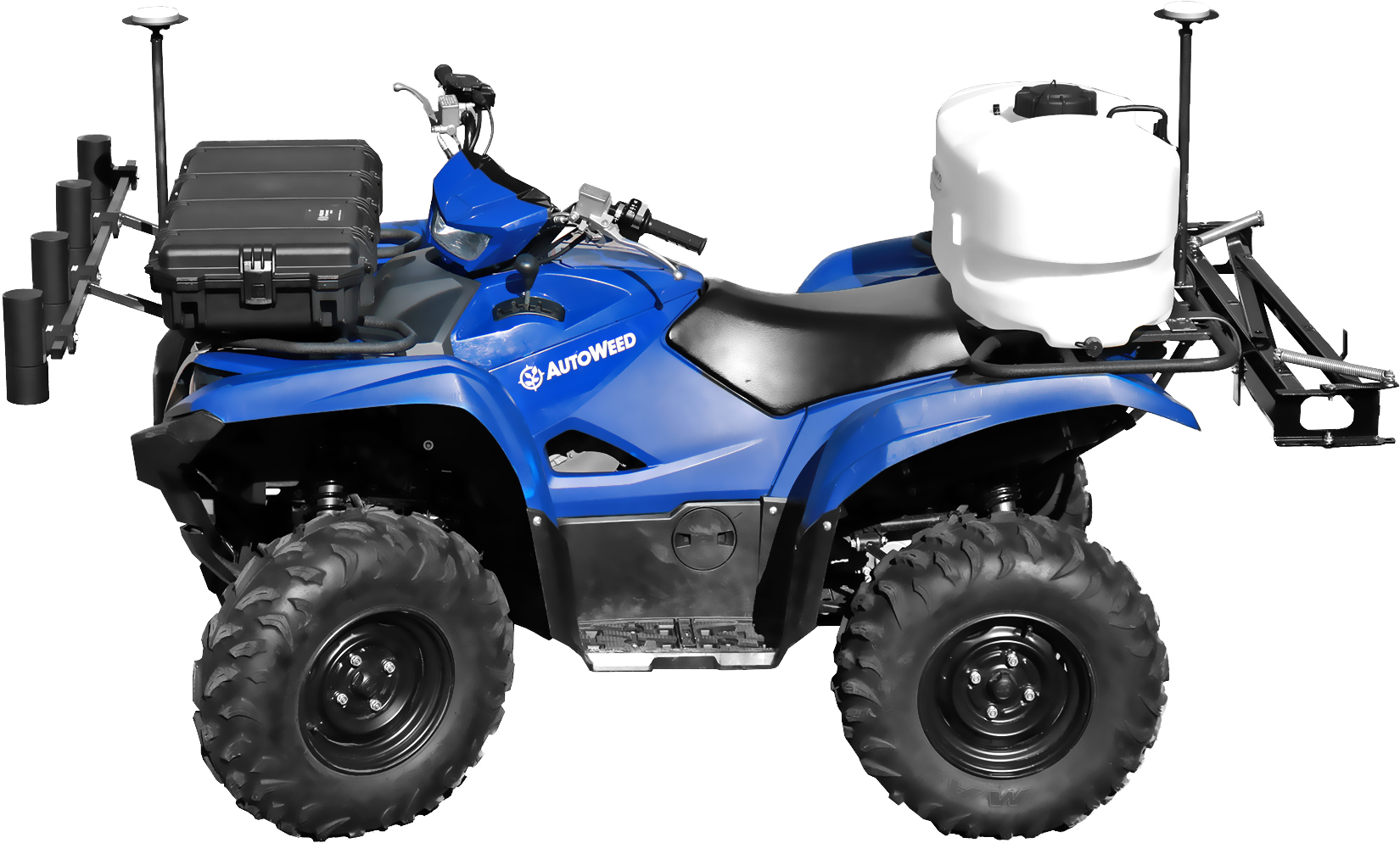}
	\caption{\emph{AutoWeed}: A prototype weed control robot for selective foliar spot spraying in the Australian rangelands.\label{AutoWeed}}
\end{figure}

Oftentimes image processing frameworks fail in real world application because they are hamstrung by unforeseen errors during the first and most important step in the framework: image acquisition \cite{Moeslund-ImageProcessing-2012}. The images acquired must match the target application as closely as possible for real world success. Our goal is to use the collected dataset to train a ground-based weed control robot; therefore, we must tailor the dataset to match this application. Our prototype ground-based weed control robot, \emph{AutoWeed} (pictured in Figure \ref{AutoWeed}),  incorporates high-resolution cameras and fast acting solenoid sprayers to perform selective spot spraying of identified weed targets. Boundary conditions in the prototype which affect the design of the optical system and subsequent classification models include:

\begin{itemize}
	\setlength\itemsep{0em}
	\item The height from the camera lens to the ground was set to 1 m in order to allow the solution to target in-fallow weed regrowth for a variety of weed species. Similarly, the ground clearance underneath the robotic vehicle is 288 mm. Weed targets will rarely exceed this height, therefore the optical system should have a depth of field of approximately 288 mm from the ground up.
	\item No external shading or lighting is to be used for the optical system, so as not to limit the vehicle's maneuverability. The camera and lens must be chosen and utilised to adequately capture dynamic lighting in the scene.
	\item The ideal vehicle speed while spraying is $10 \text{ km/hr}$. The field of view of the optical system is $450 \times 280$ mm. This gives the system approximately $100$ ms per image (or 10 fps) to detect a weed target before a new image is captured and ready for processing. This vehicle speed also requires a fast shutter speed to resolve images without motion blur.
	\item The system will operate under harsh environmental conditions. Therefore we look exclusively at machine vision cameras and lenses to provide the robust mechanical specification required here. Similarly, it is beneficial for external camera and lens parameters (such as focus, iris, zoom) to be fixed.
\end{itemize}

\begin{figure}[h]
	\centering
	\includegraphics[width=0.76\textwidth]{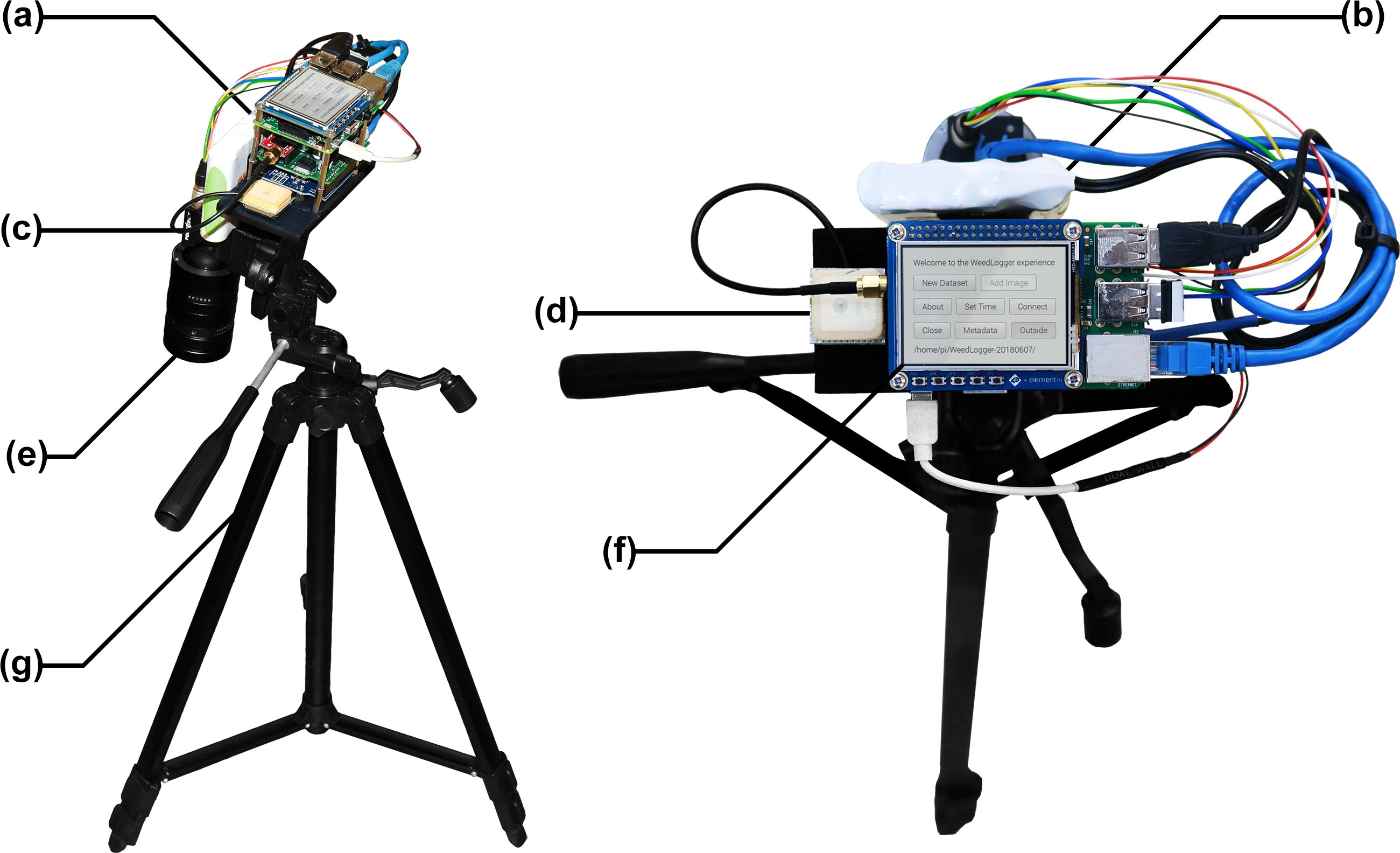}
	\caption{The \emph{WeedLogger} field instrument was developed to facilitate precise, consistent and fast collection of time and GPS stamped images. The instrument consists of: (a) a Raspberry Pi 3, Arduino Uno and custom electronics shield, (b) a rechargeable lithium-ion battery pack, (c) a FLIR Blackfly 23S6C Gigabit Ethernet high-resolution colour camera, (d) a SkyTraq Venus638FLPx GPS receiver and V.Torch VTGPSIA-3 GPS antenna, (e) a Fujinon CF25HA-1 machine vision lens, (f) a 4D Systems Raspberry Pi touchscreen display module and (g) an Inca i330G light-weight tripod. \label{WeedLogger}}
\end{figure}

A data logging instrument was developed to: (1) photograph images with the same optical system as the target robotic platform, (2) ensure consistent image acquisition and (3) accelerate the image collection process. The \emph{WeedLogger} (pictured in Figure \ref{WeedLogger}) consists of a Raspberry Pi, high resolution camera, machine vision lens and a GPS receiver. The FLIR Blackfly 23S6C Gigabit Ethernet high-resolution colour camera was chosen for this design. Its large (1920 x 1200 px) and high dynamic range (73.90 dB) image sensor affords robust imaging of a wide field of view in our highly contrasted scene. The 25mm fixed focal length Fujinon CF25HA-1 machine vision lens was paired with the $1/1.2''$ image sensor to provide a 254 mm depth of field focused to a working distance just above the ground with an aperture of $f/8$. At a working distance of 1 m, this optical system provides a 450 mm $\times$ 280 mm field of view for one image. This translates to just over 4 px per mm resolution, which has proven sufficient for leaf texture recognition in past work \cite{HoG2015}.

The lens' mode of operation was designed to resolve detail in the shadows and highlights of high dynamic range scenes without motion blur while moving at high speeds. This was achieved by selecting an aperture size of $f/8$ to allow some sunlight in; while simultaneously restricting the shutter speed to less than 0.05 ms. The automatic exposure and automatic white balance algorithms within FLIR's FlyCapture Software Development Kit were utilised to achieve acceptable imaging without the need for manual tuning between different sites. Nevertheless, colour variations will occur in the images due to changing light conditions in the natural environment throughout the day. Rather than accounting for this directly, our preference is to capture this variability in the training set for associated machine learning algorithms. Finally, the touchscreen interface allowed for in-field labelling of geo-mapped images. GPS data was collected automatically using a SkyTraq Venus638FLPx GPS receiver, V.Torch VTGPSIA-3 GPS antenna, an Arduino Uno and custom electronics shield. The GPS data was used exclusively to track progress during the dataset collection process. GPS data has not been used in the development of our classification models.

\subsection*{Dataset Collection}

When designing models or algorithms for learning features, our goal is to separate the factors of variation that explain the observed data \cite{Goodfellow-et-al-2016}. The \emph{depth} of a deep learning model conceptually refers to said model's layer count and parameter complexity. Typically, the more confounding factors of variability in the dataset, the deeper and more complex the model required to achieve acceptable performance \cite{Goodfellow-et-al-2016}. Despite our efforts to mitigate inter-scene variance of photographed images in the design of the optical system; scene and target variability will persist in our target application. Thus, a major design consideration in the construction of this dataset is to capture images that reflect the full range of scene and target variability in our target application. Hence we have chosen to abide several factors of variation, namely: illumination, rotation, scale, focus, occlusion, dynamic backgrounds; as well as geographical and seasonal variation in plant life.

Illumination will vary throughout the day with changing sunlight and canopy cover creating highly dynamic range scenes with bright reflectance and dark shadows. Rotation and scale of the target weed species will vary as they are being photographed in situ with unknown size and orientation. The distance of photographed weed species to the camera are also variable. Therefore, the fixed focal region of the camera will cause some targets to be blurred and out of focus. Fortunately, motion blur is mitigated by operating with an extremely fast shutter speed. Perhaps the most variability in the dataset is due to complex and dynamic target backgrounds. The locations subject to dense weed infestations are also inhabited by immeasurable counts of other native species. As we are unable to curate a dataset of all plant life, we must concede to labelling all other non-target plant life as \emph{negative} samples; along with all non-target background imagery. Unfortunately, this creates a highly variable class in the dataset that will be difficult to consistently classify.

In addition to complex backgrounds, target foreground objects can be unexpectedly occluded from view by interfering objects; more often other neighbouring flora. This is yet another unavoidable factor of variation. Finally, the dataset must account for seasonal variation in our target weed species. This means that a single class of weed species will include photographs of the weed with and without flowers and fruits and in varying health condition; which can affect foliage colour, strength of features and other visible anomalies.

\begin{figure}[h!]
	\centering
	\includegraphics[width=0.91\textwidth]{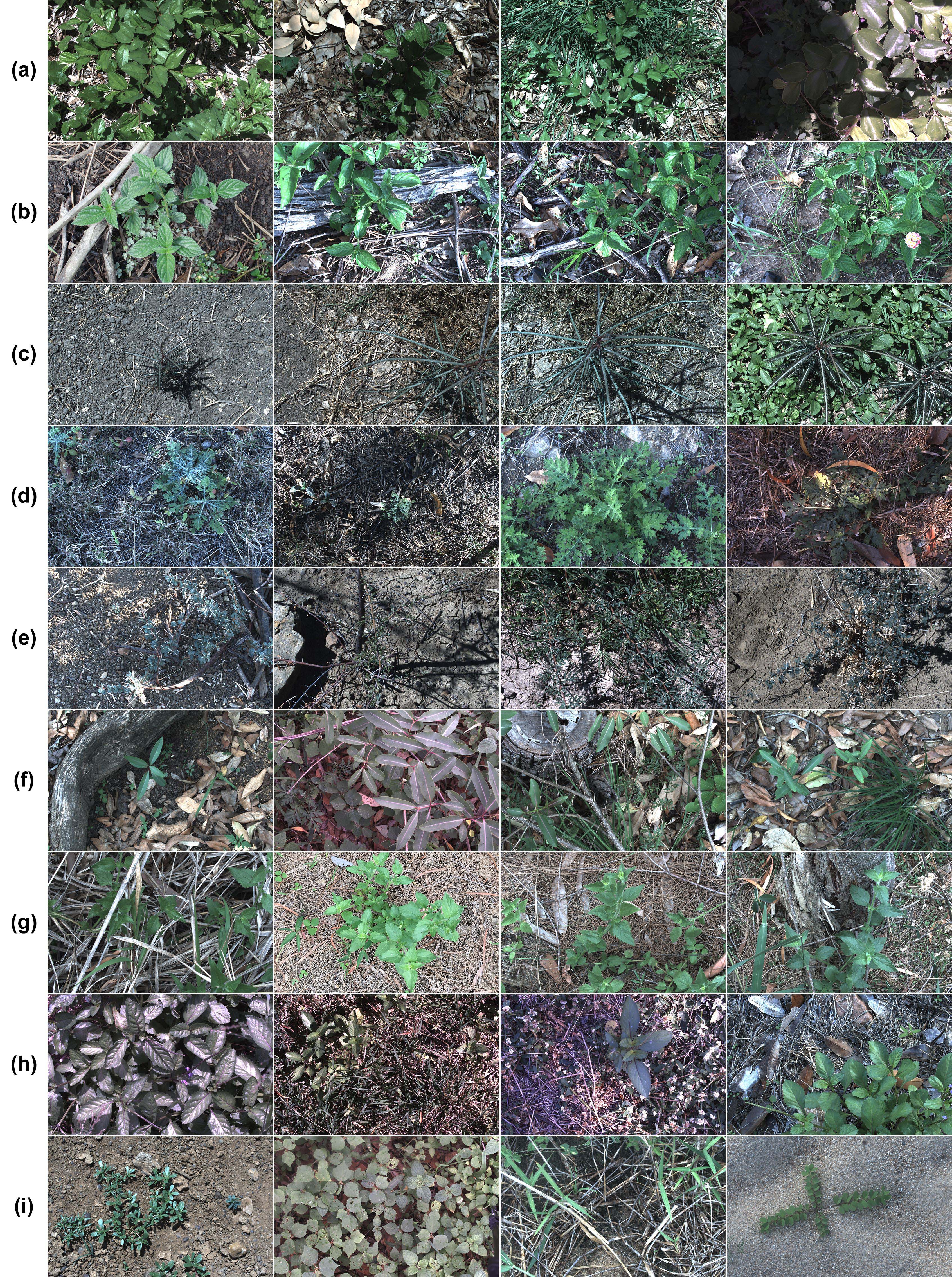}
	\caption{Sample images from each class of the \emph{DeepWeeds} dataset, namely: (a) Chinee apple, (b) Lantana, (c) Parkinsonia, (d) Parthenium, (e) Prickly acacia, (f) Rubber vine, (g) Siam weed, (h) Snake weed and (i) Negatives.\label{ExampleImages}}
\end{figure}

Two primary goals were established to achieve the required variability and generality of the dataset. First, collect at least 1,000 images of each target species. Second, attain a $50:50$ split of positive to negative class images from each location. The first goal is a necessity when training high-complexity CNNs which require large labelled datasets. The second goal helps to prevent over-fitting of developed models to scene level image features by ensuring targets are identified from their native backgrounds. Finally, the dataset required expert analysis to label each image as to whether it contains a target weed species or not. The rigidity of this collection process will ensure that the accuracy and robustness of all learning models developed to classify from it, will be upheld when applied in the field. Figure \ref{ExampleImages} illustrates a sampling of images from each class in the dataset. From this, the complexity of the learning problem is apparent due to the inherent variation within classes of the dataset.

\subsection*{Deep Learning}

The second goal of this study was to establish the baseline accuracy expected from modern deep learning CNNs; the state-of-the-art computer vision solution. An emphasis was given to off-the-shelf CNNs, which could be easily trained and deployed to facilitate wider use of the presented dataset. To that effect, the high-level neural network Application Programming Interface (API), Keras \cite{chollet2015keras}, was utilised; together with the machine learning framework, TensorFlow \cite{tensorflow2015-whitepaper}. Two popular CNN models were chosen for implementation based on their strong performance on highly variable datasets and their availability in the Keras and TensorFlow backend. The first model chosen, Inception-v3 \cite{InceptionV3} (the third improvement upon the GoogLeNet \cite{GoogLeNet2015} Inception architecture) was the winner of ILSVRC 2014 \cite{ILSVRC15}; and the second, ResNet-50 \cite{ResNet}, was the winner of ILSVRC 2015 \cite{ILSVRC15}. This annual competition sees participant models recognise 1,000 different ImageNet object classes from over 500,000 images. Other winners of the ILSVRC were considered at the time of writing. The comparable 2014 winner, VGGNets \cite{Simonyan14c} was dismissed due to having 144 million training parameters compared to the more manageable 21.8 million parameters of Inception-v3. The 2016 ILSVRC winner, GBD-Net \cite{GBDNet2016}, was not considered because it requires within-image per-pixel labels, which are not yet available for the \emph{DeepWeeds} dataset. Furthermore, the 2017 competition winner, Squeeze-and-Excitation (SE) networks \cite{ImageNet2017ClassWinner} was not considered because the SE augmented models are not readily available in the Keras and TensorFlow backend.

The models considered henceforth, Inception-v3 \cite{InceptionV3} and ResNet-50 \cite{ResNet}, are both available in Keras with pre-trained weights in the TensorFlow backend. The models were trained to recognise the 1,000 different ImageNet \cite{ILSVRC15} object classes. Their original ImageNet-trained architectures were slightly modified to classify the nine \emph{DeepWeeds} species classes. This was achieved by replacing their last fully connected 1,000 neuron layer with a 9 neuron fully-connected layer. Specifically, let $b$ denote the number of images per training batch, and $r$ and $c$ denote the number of pixel rows and columns, respectively. Each network then accepted a $b \times r \times c \times 3$ input matrix, where $3$ was the number of image colour channels. After removing the fully connected 1,000 neuron ImageNet output layer, both network outputs had the $b \times r_{\alpha} \times c_{\alpha} \times f_{\alpha}$ shape, where $\alpha$ denotes the specific network, and $f_{\alpha}$ was the number of extracted features for each $r_{\alpha} \times c_{\alpha}$ spatial location. Identical to Inception-v3, the spatial average pooling layer was used to convert the fully-convolutional $b \times r_{\alpha} \times c_{\alpha} \times f_{\alpha}$ output to the $b \times f_{\alpha}$  shape, which was then densely connected to the final $b \times 9$ weed classification layer. With 32 images per batch and $224 \times 224 \times 3$ sized input images, the deployed global average pooling was effectively identical to the $7 \times 7$ average pooling in the ResNet-50 model.

The nature of this classification task and the \emph{DeepWeeds} dataset allows for multiple weed species to be present in each image. Therefore, a sigmoid activation function was used for each weed-specific neuron in the output layer. This allowed an output of probabilities for each class to identify the likelihood that an image belonged to each class. An image was classified as one of the target weeds if that class' sigmoid-activated neuron probability was the greatest amongst all others and its likelihood was greater than $1/9=11.\overline{1}\%$ (i.e. a random guess). The random guess threshold was implemented to overcome the immense variation in the negative \emph{DeepWeeds} class, which causes its target probability to be less strongly weighted towards specific image features than the eight positive classes -- whose images are more consistent.

All 17,509 labelled images from \emph{DeepWeeds} were partitioned into 60\%-20\%-20\% splits of training, validation and testing subsets for $k$-fold cross validation with $k=5$. Stratified random partitioning was performed to ensure even distribution of the classes within each subset, except for the negative class which is much larger. The 60\% random split constitutes the training subset, while 20\% were used as the validation subset to monitor the training process and minimise over-fitting. The random splits for each fold were controlled by a random seed such that the individual split could be reproduced as required. The remaining 20\% of images were reserved for testing and were not used in any way during the training process. Before training, each model was loaded with the corresponding weights pre-trained on ImageNet. The weights of the 9-neuron fully-connected layer ({\em i.e.} the output layer) were initialised by the uniform random distribution as per Glorot et al\cite{pmlr-v9-glorot10a}. The standard {\em binary cross-entropy} loss function was used for training.

To overcome the highly variable nature of the target weed classification application, a series of augmentations were performed on both the training and validation image subsets to account for variations in rotation, scale, colour, illumination and perspective. Image augmentation was performed using the Open Source Computer Vision Library (OpenCV) and its Python wrapper. All images were first resized to $256 \times 256 $ pixels in size and randomly augmented for each epoch of training, {\em i.e.} one pass through all available training and validation images. Each image was also randomly rotated in the range of  $[-360, +360]$ degrees. Then, each image was randomly scaled both vertically and horizontally in the range of $[0.5, 1]$. Each colour channel was randomly shifted within the range of $\pm 25$ ({\em i.e.} approximately $\pm 10\%$ of the maximum available 8-bit colour encoding range $[0,255]$). To account for illumination variance, pixel intensity was randomly shifted within the $[-25, +25]$ range, shifting all colour channels uniformly. In addition, pixel intensity was randomly scaled within the $[0.75, 1.25]$ range. Random perspective transformations were applied to each image to simulate a large variation of viewing distances and angles. Finally, the images were flipped horizontally with a 50\% probability and then cropped to retain the $224 \times 224$ pixels required for each architecture's input layer. With all nine classes of the \emph{DeepWeeds} dataset, the ResNet-50 and Inception-v3 models contained approximately 23.5 million and 21.8 million trainable weights, respectively. Without this extensive augmentation, both CNN networks drastically over-fitted the available images by memorising the training subsets.

The Keras implementation of Adam\cite{Adamax14}, a first-order gradient-based method for stochastic optimisation, was used for the training of both models. The initial learning-rate ($lr$) was set to $lr=1\times 10^{-4}$. It was then successively halved every time the \emph{validation loss} did not decrease after 16 epochs. Note that the validation loss refers to the classification error computed on the validation subset of images. The training was performed in batches of 32 images, and aborted if the validation loss did not decrease after 32 epochs. While training, the model with the smallest running validation loss was continuously saved, in order to re-start the training after an abortion. In such cases, training was repeated with the initial learning rate $lr=0.5 \times 10^{-4}$.

\begin{figure}
	\centering
	\includegraphics[width=0.95\textwidth]{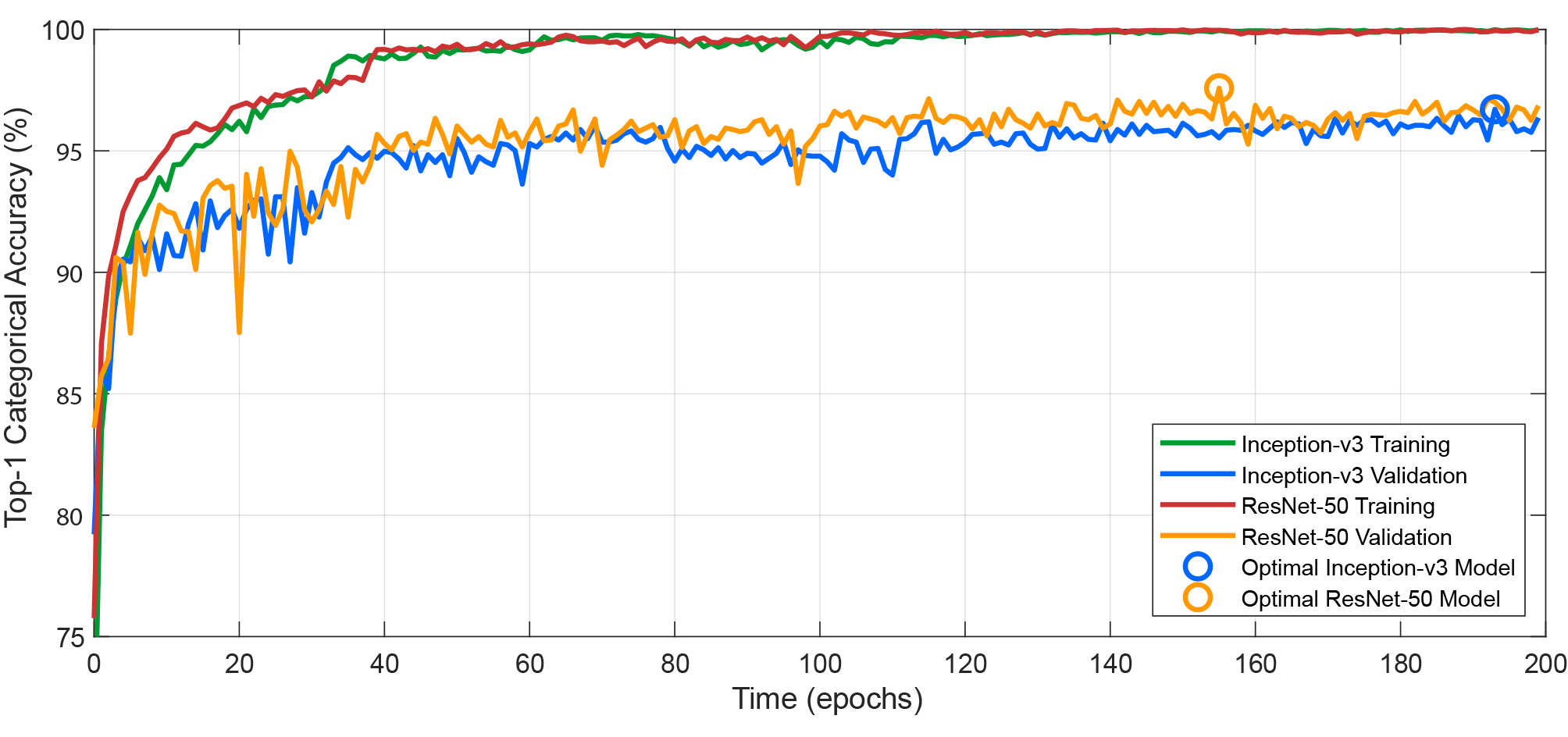}
	\caption{Visualisation of the learning process for a single cross validated fold where the training and validation accuracy for Inception-v3 and ResNet-50 improve after successive epochs; beginning to plateau after 100 epochs. Optimal validation accuracies of 96.7\% and 97.6\% were achieved at epochs 193 and 155 for Inception-v3 and ResNet-50, respectively.\label{ExampleTrainingFigure}}
\end{figure}

Figure \ref{ExampleTrainingFigure} illustrates the learning process for both the Inception-v3 and ResNet-50 models on the \emph{DeepWeeds} training and validation subsets. It can be seen that both methods plateaued in accuracy after roughly 100 epochs. It took an average of 13 hours to train a single model on an NVIDIA GTX 1080Ti Graphical Processing Unit (GPU), where ResNet-50 and Inception-v3 consumed comparable training times.

\section*{Results}

\subsection*{The \emph{DeepWeeds} Dataset}

From June 2017 to March 2018, images were collected from sites across northern Australia using the \emph{WeedLogger} in-field instrument. The result is \emph{DeepWeeds}, a large multiclass dataset comprising 17,509 images of eight different weed species and various off-target (or negative) plant life native to Australia.

\begin{table*}[h]
	\centering
	\caption{The distribution of \emph{DeepWeeds} images by weed species (row) and location (column).\label{DeepWeedsCounts}}
	\begin{tabular}{|c|cccccccc|c|}
		\hline
		&\itshape \parbox{1.2cm}{Black River} & \itshape \parbox{1.2cm}{Charters Towers} & \itshape \parbox{1.2cm}{Cluden} & \itshape \parbox{1.2cm}{Douglas} & \itshape \parbox{1.2cm}{Hervey Range} & \itshape \parbox{1.2cm}{Kelso} & \itshape \parbox{1.2cm}{McKinlay} & \itshape \parbox{1.2cm}{Paluma} & Total\\
		\hline
		\itshape Chinee apple & 0 & 0 & 0 & 718 & 340 & 20 & 0 & 47 & 1125\\
		\itshape Lantana & 0 & 0 & 0 & 9 & 0 & 0 & 0 & 1055 & 1064\\
		\itshape Parkinsonia & 0 & 0 & 1031 & 0 & 0 & 0 & 0 & 0 & 1031\\
		\itshape Parthenium & 0 & 246 & 0 & 0 & 0 & 776 & 0 & 0 & 1022\\
		\itshape Prickly acacia & 0 & 0 & 132 & 1 & 0 & 0 & 929 & 0 & 1062\\
		\itshape Rubber vine & 0 & 188 & 1 & 815 & 0 & 5 & 0 & 0 & 1009\\
		\itshape Siam weed & 1072 & 0 & 0 & 0 & 0 & 0 & 0 & 2 & 1074\\
		\itshape Snake weed & 10 & 0 & 0 & 928 & 1 & 34 & 0 & 43 & 1016\\
		\itshape Negatives & 1200 & 605 & 1234 & 2606 & 471 & 893 & 943 & 1154 & 9106\\
		\hline
		Total & 2282 & 1039 & 2398 & 5077 & 812 & 1728 & 1872 & 2301 & 17509\\
		\hline
	\end{tabular}
\end{table*}

\begin{figure}[h!]
	\centering
	\includegraphics[width=0.93\textwidth]{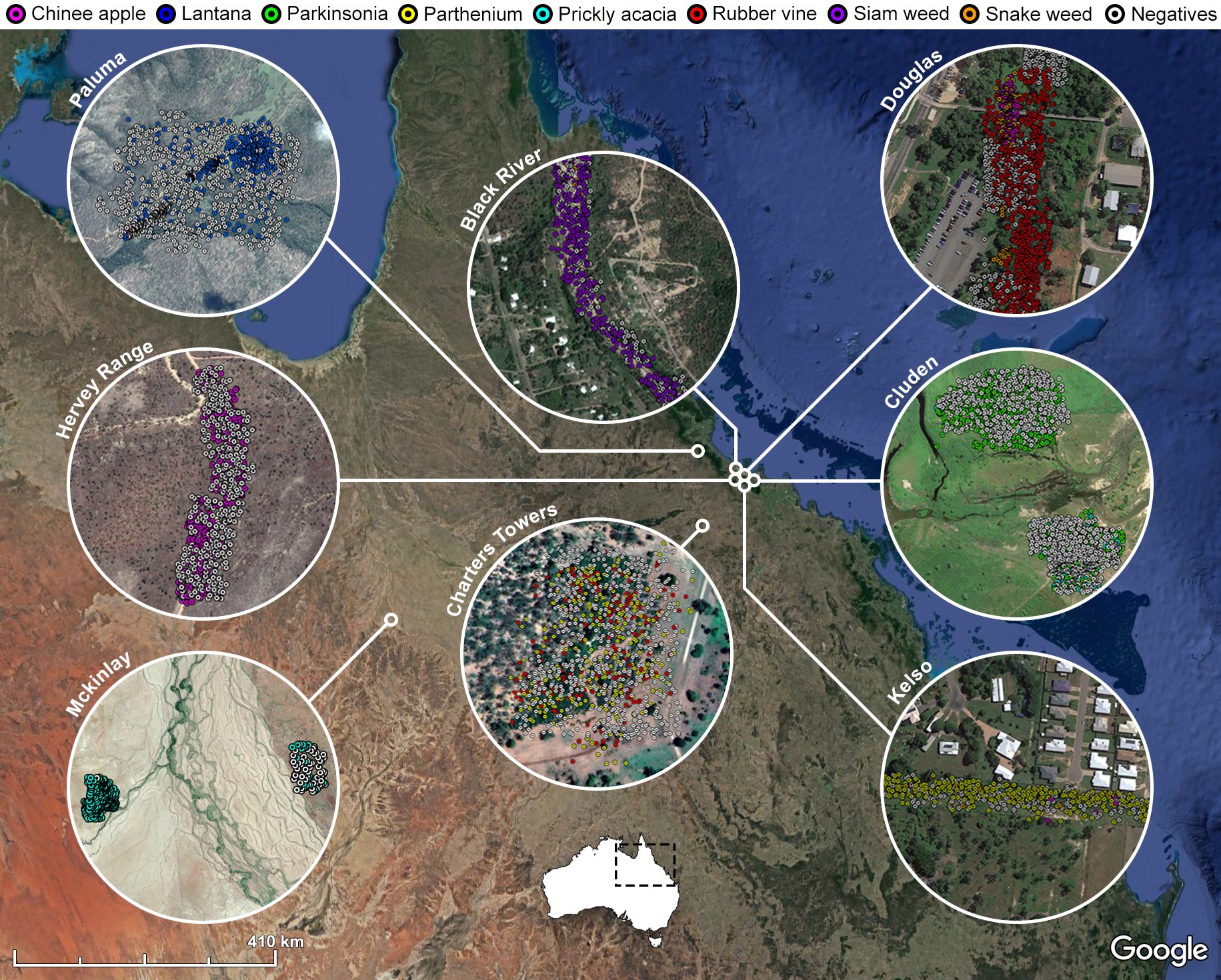}
	\caption{The geographical distribution of \emph{DeepWeeds} images across northern Australia (Data: Google, SIO, NOAA, U.S. Navy, NGA, GEBCO; Image \textcopyright \space 2018 Landsat / Copernicus; Image \textcopyright \space 2018 DigitalGlobe; Image \textcopyright \space 2018 CNES / Airbus). \label{DeepWeedsMap}}
\end{figure}

Table \ref{DeepWeedsCounts} shows the quantitative distribution of images, sorted by weed species and location. Over 1,000 images were collected of each weed species, totaling over 8,000 images of positive species classes. Images of neighbouring flora and backgrounds that did not contain the weed species of interest were collated into a single ``negative'' class. To balance any scene bias, an even split of positive and negative samples were collected from each location. This balance can be observed in Figure \ref{DeepWeedsMap} which maps the geographical distribution of the images.

The breadth of the dataset is also apparent from Figure \ref{DeepWeedsMap}, which spans several collection sites across northern Australia. Each collection site was identified to contain large infestations of specific target species, including: siam weed from \emph{Black River} ($19\degree 13' 44''$ S, $146\degree 37' 45''$ E), rubber vine from \emph{Charters Towers} ($20\degree 04' 44''$ S, $146\degree 10' 55''$ E), parkinsonia from \emph{Cluden} ($19\degree 19' 02''$ S, $146\degree 51' 02''$ E), snake weed from \emph{Douglas} ($19\degree 19' 29''$ S, $146\degree 45' 44''$ E), chinee apple from \emph{Hervey Range} ($19\degree 19' 35''$ S, $146\degree 38' 50''$ E), parthenium from \emph{Kelso} ($19\degree 22' 38''$ S, $146\degree 43' 05''$ E), prickly acacia from \emph{McKinlay} ($21\degree 20' 21''$ S, $141\degree 31' 27''$ E) and lantana from \emph{Paluma} ($18\degree 57' 23''$ S, $146\degree 02' 17''$ E).

\subsection*{Deep Learning Performance}
The \emph{DeepWeeds} dataset was classified with the ResNet-50 and Inception-v3 CNN models to establish a baseline level of performance for future comparison. Results are tabulated below. Table \ref{DLPrecision} documents a variety of metrics evaluating the average classification performance across all five cross validated folds for both Inception-v3 and ResNet-50. We specifically look at the class weighted top-1 accuracy (\%), precision (\%) and false positive rate (\%) averaging across the five folds. Meanwhile, Table \ref{DLConfusion} provides the combined within-class confusion matrix across the five cross validated folds for ResNet-50, which was found to perform slightly better than Inception-v3.

Table \ref{DLPrecision} illustrates that Inception-v3 and ResNet-50 achieved an average top-1 classification accuracy of 95.1\% and 95.7\%, respectively. This is a very strong baseline result. ResNet-50 also outperformed Inception-v3 in terms of both precision and false positive rate. The slightly better performance of the ResNet-50 model may be attributed to its slightly higher complexity, constituting 23.5 million trainable weights, as opposed to 21.8 million for Inception-v3. This added complexity allows for a broader optimisation space within which the ResNet-50 model can learn.

The classification accuracy is seen to vary from species to species for both models. The strongest performance for the ResNet-50 model was achieved on the negative class with 97.6\% average top-1 accuracy. This is a strong and surprising result considering the immense variation in this class. Other weed species with strong performance are parkinsonia at 97.2\%, siam weed at 96.5\%, parthenium at 95.8\%, lantana at 95.0\% and prickly acacia at 95.5\%. The weed species with the lowest classification accuracy were rubber vine at 92.5\%, snake weed at 88.8\% and chinee apple at 88.5\%. We suspect the relatively low performance on these weed species may be due to the weed having less unique visible features to train on.

Observing the precision statistics of Table \ref{DLPrecision} for the ResNet-50 model, it can be seen that the confidence in the model's positive predictions are quite high (greater than 90\% for all species). The three species with the strongest positive predictive value are rubber vine at 99.1\%, parkinsonia at 97.9\% and parthenium at 96.7\%. The negative class prediction is also of high confidence with a precision score of 96.7\%. Comparatively, chinee apple, lantana, prickly acacia and snake weed demonstrate the lowest relative confidence in their positive predictions ranging from 91.0\% to 93.0\%. This may be attributed to the high confusion rates between these specific weed species due to similar image features.

The average false positive rate for both models is around 2\%, with the individual false positive rates for each weed species sitting well below 1\% (as shown in Table \ref{DLPrecision}). This strong result indicates that our models are likely to result in minimal off-target damage when deployed in the field. However, the negative class has a relatively high false positive rate of 3.77\% and 3.59\% for Inception-v3 and ResNet-50, respectively. Because this class contains non-target plant life, this tells us that over 3\% of our weed targets are being falsely classified as belonging to the negative class. We reason this is due to the large variation of plant life in the negative class across the eight rural locations for the dataset. Consequently, when deploying such models in the field it is beneficial to train location specific models to further minimise the number of misclassified targets.

\begin{table}[h]
	\setlength\aboverulesep{0pt}
	\setlength\belowrulesep{0pt}
	\centering
	\caption{The average test classification accuracy, recall rate, precision and false positive rate across all five cross validated folds for both Inception-v3 and ResNet-50. The statistic from the best performing network is emboldened for each species. Equations for the computation of each metric are provided below. \label{DLPrecision}}
    \begin{tabular}{|l|c|c|c|c|c|c|}
		\toprule
		\multicolumn{1}{|c|}{\multirow{2}[4]{*}{Species}} & \multicolumn{2}{c|}{Top-1 accuracy (\%)} & \multicolumn{2}{c|}{Precision (\%)} & \multicolumn{2}{c|}{False positive rate (\%)} \\
		\cmidrule{2-7}          & Inception-v3 & ResNet-50 & Inception-v3 & ResNet-50 & Inception-v3 & ResNet-50 \\
		\midrule
		\textit{Chinee Apple} & 85.3  & \textbf{88.5} & \textbf{92.7} & 91.0  & \textbf{0.48} & 0.61 \\
		\textit{Lantana} & 94.4  & \textbf{95.0} & 90.9  & \textbf{91.7} & 0.62  & \textbf{0.55} \\
		\textit{Parkinsonia} & 96.8  & \textbf{97.2} & 95.6  & \textbf{97.9} & 0.29  & \textbf{0.13} \\
		\textit{Parthenium} & 94.9  & \textbf{95.8} & 95.8  & \textbf{96.7} & 0.26  & \textbf{0.21} \\
		\textit{Prickly Acacia} & 92.8  & \textbf{95.5} & \textbf{93.4} & 93.0  & \textbf{0.43} & 0.46 \\
		\textit{Rubber Vine} & \textbf{93.1} & 92.5  & \textbf{99.2} & 99.1  & \textbf{0.05} & 0.05 \\
		\textit{Siam Weed} & \textbf{97.6} & 96.5  & 94.4  & \textbf{97.2} & 0.38  & \textbf{0.18} \\
		\textit{Snake Weed} & 88.0  & \textbf{88.8} & 86.9  & \textbf{90.9} & 0.82  & 0.55 \\
		\textit{Negatives} & 97.2  & \textbf{97.6} & 96.5  & \textbf{96.7} & 3.77  & 3.59 \\
		\midrule
		Weighted average & 95.1  & \textbf{95.7} & 95.1  & \textbf{95.7} & 2.16  & \textbf{2.04} \\
		\bottomrule
	\end{tabular}%
\end{table}

\begin{table*}[h]
	\setlength\aboverulesep{0pt}
	\setlength\belowrulesep{0pt}
	\centering
	\caption{The confusion matrix (\%) achieved by the ResNet-50 model on the test subsets for the five cross validated folds.\label{DLConfusion}}
	\begin{tabular}{|c|ccccccccc|}
		\toprule
		& \multicolumn{1}{p{3em}}{\textit{Chinee apple}} & \multicolumn{1}{p{3em}}{\textit{Lantana}} & \multicolumn{1}{p{4.5em}}{\textit{Parkinsonia}} & \multicolumn{1}{p{4.5em}}{\textit{Parthenium}} & \multicolumn{1}{p{3em}}{\textit{Prickly acacia}} & \multicolumn{1}{p{3em}}{\textit{Rubber vine}} & \multicolumn{1}{p{3em}}{\textit{Siam\newline{}weed}} & \multicolumn{1}{p{3em}}{\textit{Snake weed}} & \multicolumn{1}{p{4.5em}|}{\textit{Negatives}} \\
		\midrule
		\textit{Chinee apple} & \textbf{88.5} & 1.78  & 0.00  & 0.44  & 0.18  & 0.18  & 0.27  & 3.37  & 5.33 \\
		\textit{Lantana} & 0.56  & \textbf{95.0} & 0.00  & 0.00  & 0.00  & 0.09  & 0.28  & 0.94  & 3.10 \\
		\textit{Parkinsonia} & 0.10  & 0.00  & \textbf{97.2} & 0.10  & 1.26  & 0.00  & 0.00  & 0.00  & 1.36 \\
		\textit{Parthenium} & 0.10  & 0.20  & 0.10  & \textbf{95.8} & 0.88  & 0.10  & 0.00  & 0.29  & 2.54 \\
		\textit{Prickly acacia} & 0.00  & 0.00  & 0.56  & 0.66  & \textbf{95.5} & 0.00  & 0.00  & 0.09  & 3.20 \\
		\textit{Rubber vine} & 0.79  & 0.50  & 0.10  & 0.10  & 0.00  & \textbf{92.5} & 0.20  & 0.40  & 5.45 \\
		\textit{Siam weed} & 0.00  & 0.19  & 0.00  & 0.00  & 0.00  & 0.00  & \textbf{96.5} & 0.09  & 3.26 \\
		\textit{Snake weed} & 4.13  & 1.77  & 0.00  & 0.30  & 0.20  & 0.10  & 0.30  & \textbf{88.8} & 4.43 \\
		\textit{Negatives} & 0.46  & 0.48  & 0.14  & 0.20  & 0.55  & 0.03  & 0.21  & 0.37  & \textbf{97.6} \\
		\bottomrule
	\end{tabular}%
\end{table*}

\newpage

\begin{figure}[h!]
	\centering
	\includegraphics[width=0.95\textwidth]{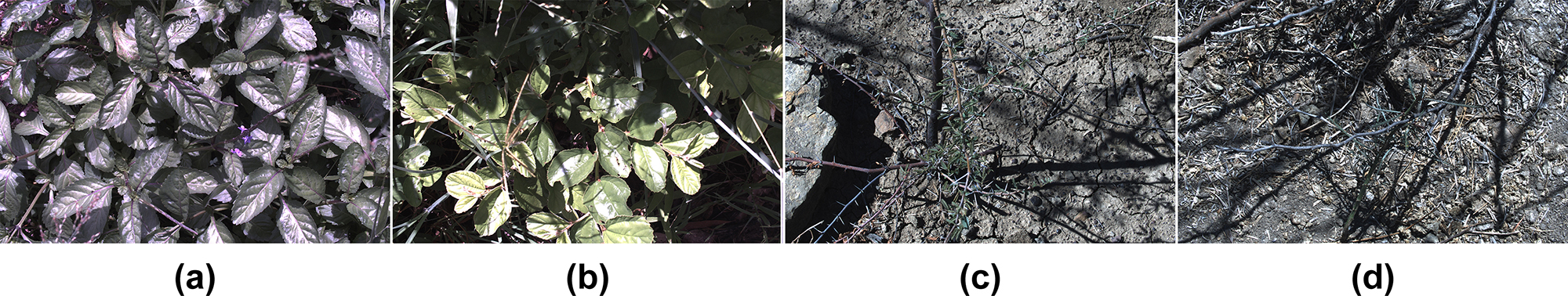}
	\caption{Example images highlighting confusions between classes of weed species. Specifically, (a) correctly classified snake weed, (b) chinee apple falsely classified as snake weed, (c) correctly classified prickly acacia, and (d) parkinsonia falsely classified as prickly acacia. \label{ConfusionExamples}}
\end{figure}

Table \ref{DLConfusion} shows the confusion matrix resulting from combining the ResNet-50 model's performance across the five cross validated test subsets. The model confuses 3.4\% of chinee apple images with snake weed and 4.1\% vice versa. Reviewing these particular samples shows that under certain lighting conditions the leaf material of chinee apple looks strikingly similar to that of snake weed. This is illustrated in the sample misclassification of snake weed in Figure \ref{ConfusionExamples}. Furthermore, the ResNet-50 model incorrectly classifies 1.3\% of parkinsonia images as prickly acacia. Figure \ref{ConfusionExamples} also shows a sample misclassification of prickly acacia. It should be noted that parkinsonia and prickly acacia are from the same genus, and are commonly both known as \emph{prickle bush} species. In addition to their similar shape and size, they both produce thorns, yellow flowers and bean-like seed pods. This likeness is the reason for these false positives in our model.

\begin{figure}[h!]
	\centering
	\includegraphics[width=0.95\textwidth]{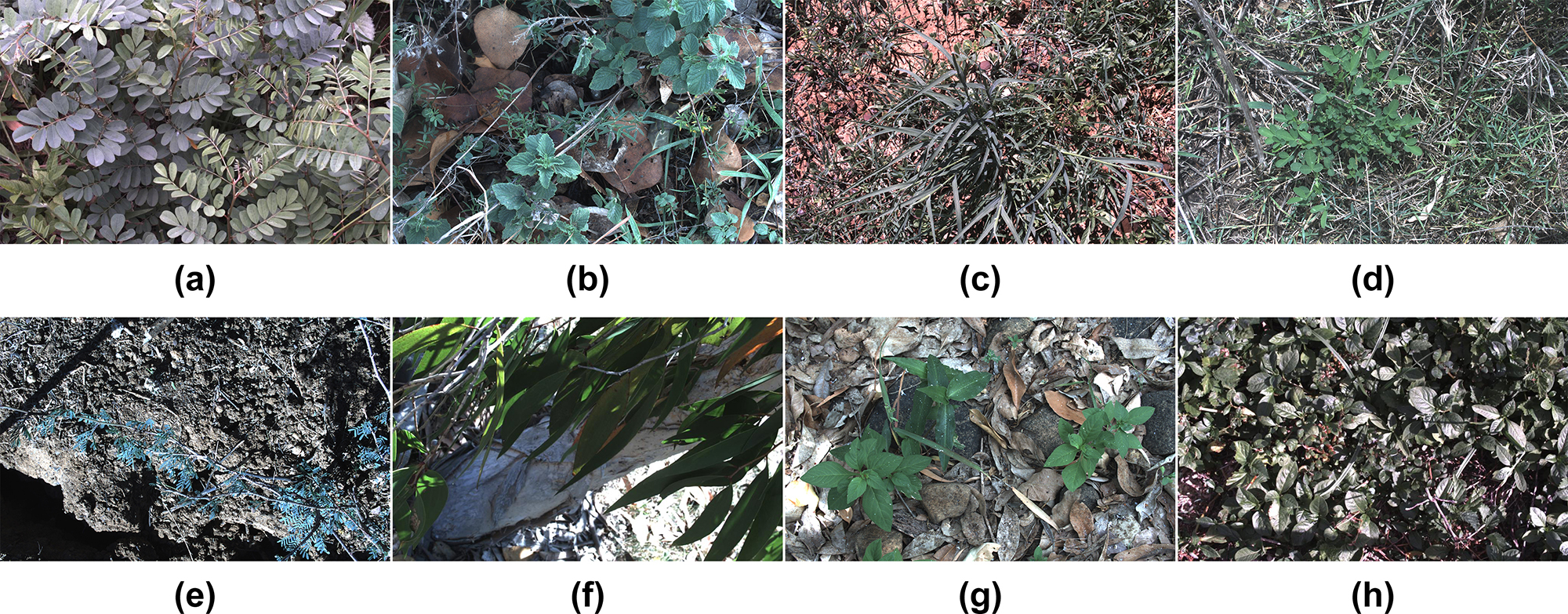}
	\caption{Example false positives where images from the negative class were falsely classified as (a) Chinee apple, (b) Lantana, (c) Parkinsonia, (d) Parthenium, (e) Prickly acacia, (f) Rubber vine, (g) Siam weed and (h) Snake weed. \label{FalsePositives}}
\end{figure}

While these minor confusions are indeed limitations of the learning model, we should highlight that chinee apple, snake weed, parkinsonia and prickly acacia are all weeds that need to be controlled. So confusing one for the other when eradicating them all is inconsequential. Of more concern are false positives where established native plant life is incorrectly classified as weeds. Consequently spraying off-target plant life may cause harm to the native ecosystem, as well as waste expensive herbicide. Figure \ref{FalsePositives} provides examples of negative plant life that has been misclassified as each of the eight target weed species. In every case, the native plant exhibits image features that closely resemble the species it has been confused with. This shows that the immense variation in plant life that makes up the negative class does introduce unavoidable false positives. Fortunately, these cases are few and far between, with false positive rates below 1\% for each species in Table \ref{DLPrecision} and less than 3\% of the negative class being falsely classified in Table \ref{DLConfusion}.

\subsection*{Real Time Inference}
As we progress toward the application of robotic weed control, we must also assess our chosen model's viability for real-time and in-field performance.
As discussed earlier, our optical system's 450 x 280 mm field of view allows approximately 100 ms of total processing time per image for the selective spot sprayer to operate at the target speed of 10 km/hr. In other words, we must be able to process images at 10 fps or more to achieve target real time performance.
The complexity of the ResNet-50 and Inception-v3 CNN models demand a dedicated and high performance GPU card. The NVIDIA Jetson TX2 is an embedded platform for power-efficient edge computing. Combining a Tegra \emph{Parker} system on a chip (SoC) with a Pascal$^{TM}$ architecture GPU, the Jetson TX2 provides an embedded platform for the inference of complex CNNs with a conservative maximum power consumption of 15 W.

To assess if our developed models are sufficient for real time robotic weed control, our best performing model, ResNet-50, was deployed on an NVIDIA Jetson TX2 compute module. Inference was performed on all images in the \emph{DeepWeeds} dataset while measuring inference time per image and the resulting achievable frame rate. Two machine learning platforms were investigated for this experiment: TensorFlow, which was also used for training; and TensorRT\cite{tensorrt}, an optimised platform for high-performance deep learning inference. Note that the time taken to perform the required pre-processing steps was also measured. These steps include loading an image and resizing it for input to the network.

\begin{table}[h!]
	\centering
	\caption{Real time inference results for a \emph{DeepWeeds} trained ResNet-50 model on an NVIDIA Jetson TX2 using the TensorFlow and TensorRT software platforms.\label{realtimeinference}}
	\begin{tabular}{|c|c|c|c|c|}
		\hline
		Platform & Inference time (ms) & Preprocessing time (ms) & Total inference time (ms) & Frame rate (FPS) \\
		\hline
		TensorFlow & $128 \pm 47$ & $51.9 \pm 4.6$ & $180 \pm 52$ & 5.55 \\
		TensorRT & $26.7 \pm 6.3$ & $26.7 \pm 1.8$ & $53.4 \pm 8.1$ & 18.7 \\
		\hline
	\end{tabular}
\end{table}

Table \ref{realtimeinference} reveals that our best performing ResNet-50 deep learning model cannot perform inference within the real time performance target of 100 ms per image using the standard TensorFlow library. However, through the use of the optimised deep learning inference engine, TensorRT, the real time performance target is met. TensorRT delivers unparalleled inference speed, taking an average of 53.4 ms to perform pre-processing and inference on a single image. The model could theoretically run at 18.7 fps, almost doubling the required target frame rate of 10 fps for our robotic weed control application.

\section*{Conclusions}

In summary, this work introduces the first, large, multiclass weed species image dataset collected entirely in situ from Australian rangelands. \emph{DeepWeeds} contains eight weed species of national significance to Australia, and spans eight geographic locations from northern Australia. We present strong baseline performance on the dataset using the Inception-v3 and ResNet-50 CNN models, that achieve an average classification performance of 95.1\% and 95.7\%, respectively. The best performing ResNet-50 model also performed inference well within the real-time requirements of the target robotic weed control application, inferring at 53.4 ms per image and 18.7 fps. These strong classification results further prove the power of deep learning for highly variable image datasets, and show that real time deployment of such complex models is viable.

We anticipate that the dataset and our classification results will inspire further research into the classification of rangeland weeds under realistic conditions. Future work in this area includes: improving the accuracy and robustness of classifying the \emph{DeepWeeds} dataset, field implementation of our learning models as the detection system for a prototype weed control robot and investigating the use of NIR spectroscopy and hyperspectral imaging for weed species classification. The great lengths taken to collect a dataset including the real life complexity of the rangeland environment should allow for strong in-field performance.

\section*{Data Availability}
The \emph{DeepWeeds} dataset and source code for this work is publicly available through the corresponding author’s GitHub repository: \url{https://github.com/AlexOlsen/DeepWeeds}.

\bibliography{DeepWeeds2018}
\newpage

\section*{Acknowledgments}
This work is funded by the Australian Government Department of Agriculture and Water Resources Control Tools and Technologies for Established Pest Animals and Weeds Programme (Grant No. 4-53KULEI).

\section*{Author contributions statement}
A.O. and D.A.K. wrote the manuscript. 
A.O., D.A.K. and B.P. conceived the experiments.
A.O., P.R., J.C.W., J.J., W.B. and B.G. collected the dataset.
P.R., R.D.W., B.P., A.O., O.K., J.W., M.R.A. and B.C. conceived the project.
B.P., P.R., R.D.W., M.R.A., O.K. and B.C. reviewed the manuscript.

\section*{Additional information}
\textbf{Competing Interests:} The authors declare no competing interests.

\end{document}